\documentclass[journal]{IEEEtran}

\usepackage{amsmath}
\usepackage{amssymb}
\usepackage{amsmath}
\usepackage{bbding}

\usepackage{multirow}
\usepackage{setspace}
\usepackage{color,subfigure}
\usepackage{mathrsfs}
\usepackage{graphics}
\usepackage{epsfig}
 \usepackage{url}

\ifCLASSINFOpdf
  
\else
 
\fi

\begin{document}
 
\title{Scale-aware Fast R-CNN for Pedestrian Detection}

\author{Jianan~Li,
        Xiaodan~Liang,
        ShengMei~Shen,
        Tingfa~Xu,
        Jiashi~Feng,
        and Shuicheng~Yan
\thanks{Jianan~Li and Tingfa~Xu are with School of Optical Engineering, Beijing Institute of Technology University, China. 
Xiaodan~Liang is from Sun Yat-Sen University, China.
ShengMei~Shen is from Panasonic R\&D Center, Singapore. 
Jiashi~Feng and Shuicheng~Yan are from Department of Electrical and Computer Engineering, National University of Singapore.
}
}

\maketitle

\begin{abstract}
In this work, we consider the problem of pedestrian detection in natural scenes. Intuitively, instances of pedestrians with different spatial scales may exhibit dramatically different features. Thus, large variance in instance scales, which  results in  undesirable large intra-category variance in features, may severely hurt the performance of modern object instance detection methods.  We argue that this issue can be substantially alleviated by the divide-and-conquer philosophy. Taking pedestrian detection as an example, we illustrate how we can leverage this philosophy to develop a Scale-Aware Fast R-CNN (SAF R-CNN) framework. The model introduces multiple built-in sub-networks which detect pedestrians with scales from disjoint ranges. Outputs from all the sub-networks are then adaptively combined to generate the final detection results that are shown to be robust to large variance in instance scales, via a gate function defined over the sizes of object proposals. Extensive evaluations on several challenging pedestrian detection datasets well demonstrate the effectiveness of the proposed SAF R-CNN. Particularly, our method achieves state-of-the-art performance on Caltech~\cite{dollar2012pedestrian}, INRIA~\cite{dalal2005histograms}, and ETH~\cite{ess2007depth}, and obtains competitive results on KITTI~\cite{geiger2012we}.
\end{abstract}

\begin{IEEEkeywords}
	Pedestrian Detection, Scale-aware, Deep Learning.
\end{IEEEkeywords}

\section{Introduction}

Pedestrian detection  aims to predict bounding boxes of all the pedestrian instances in an image. It has attracted much attention within the computer vision community in recent years~\cite{dalal2005histograms,viola2003detecting,wang2009hog,dollar2009integral,compact,dollar2014fast,zhang2014informed,felzenszwalb2010object,marin2014occlusion} as an important component for many  human-centric applications, such as self-driving vehicles, person re-identification, video surveillance and robotics~\cite{li2014deepreid,wang2014scene}.

\begin{figure}
	\begin{center}
		\includegraphics[scale=0.32]{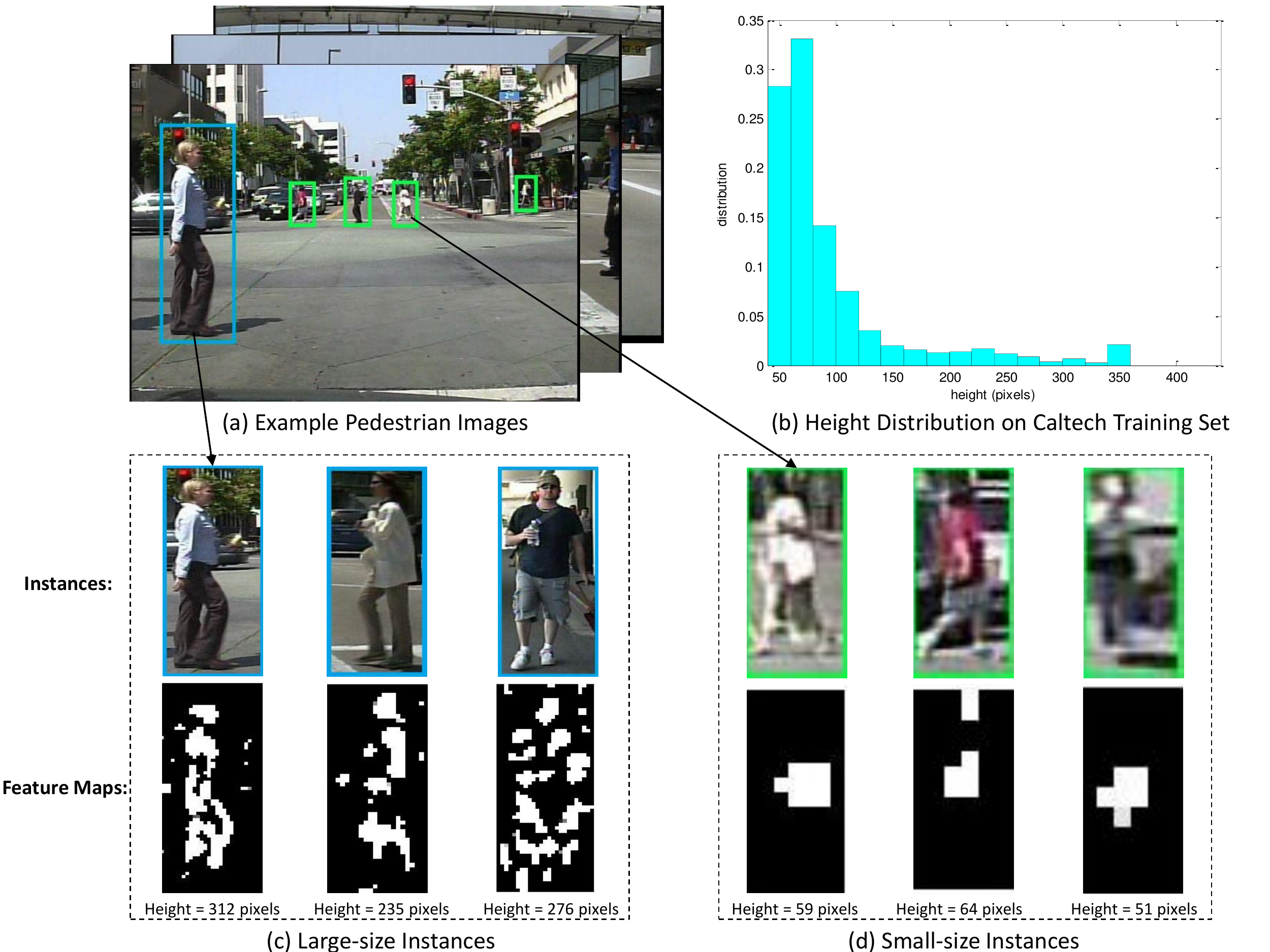}
		\caption{Illustration of the motivation of our SAF R-CNN model. (a) shows some example pedestrian images. (b) shows the distribution of pedestrians heights on the Caltech training set. One can observe that small-size (\emph{i.e.} small height) instances indeed dominate the distribution. (c) and (d) demonstrate that the visual appearance and the extracted feature maps of the large-size and small-size instances are significantly different. 
			In particular, large background clutters, obscured boundaries, body skeletons and heavy occlusion make the small-size pedestrians very difficult to detect.
		}
		\label{fig:motivation}
	\end{center}
	\vspace{-4mm}
\end{figure}

Recently, many research works~\cite{ta_cnn,compact,ouyang2013joint,sermanet2013pedestrian} have been devoted to pedestrian detection. However, they generally leave a critical issue caused by various scales\footnote{Here by ``scale'' we mean the ``size'' of the pedestrian in an image. We  use these two terms interchangeably here when no confusion is caused.} of pedestrians in an image unsolved, which is shown to considerably affect the performance of pedestrian detection in natural scenes. We provide an illustration of the motivation of the paper in Figure~\ref{fig:motivation}. Pedestrian instances in the video surveillance images (e.g., Caltech dataset~\cite{dollar2012pedestrian}) often have very small sizes. Statistically, over $60\%$ of the instances from the Caltech training set have a height smaller than $100$ pixels. Accurately localizing these small-size pedestrian instances is quite challenging due to the following difficulties. Firstly, most of the small-size instances appear with blurred boundaries and obscure appearance. It is difficult to distinguish them from the background clutters and other overlapped instances. Secondly, the large-size pedestrian instances typically exhibit dramatically different visual characteristics from the small-size ones. For instance, body skeletons of the large-size instances can provide rich information for pedestrian detection  while  skeletons of the small-size instances cannot be recognized so easily.  
Such differences can also be verified by comparing the generated feature maps for large-size and small-size pedestrians, as shown in Figure~\ref{fig:motivation}. The high feature responses for detailed body skeletons are shown for the large-size instances while only coarse feature maps are obtained for small-size instances. 

Existing works address the scale-variance problem mainly from two aspects. First, the brute-force data augmentation (e.g., multi-scaling~\cite{girshick2015fast} or resizing~\cite{girshick2014rich}) is used to improve the scale-invariance capability.  
Second, a single model~\cite{gong2014multi}\cite{xu2014scale} with multi-scale filters is employed on all instances with various sizes. However, due to the intra-class variance of large-size and small-size instances, it is difficult to handle their considerably different feature responses with a single model.  To exploit the dramatically different characteristics of instances with various scales, we adopt the divide-and-conquer philosophy to address this critical scale-variance problem. Based on this philosophy, a unified framework can comprise multiple single models, each of which specializes in detecting instances with scales of a particular range by capturing scale-specific visual patterns. 

\begin{figure*}
	\begin{center}
		\includegraphics[scale=0.52]{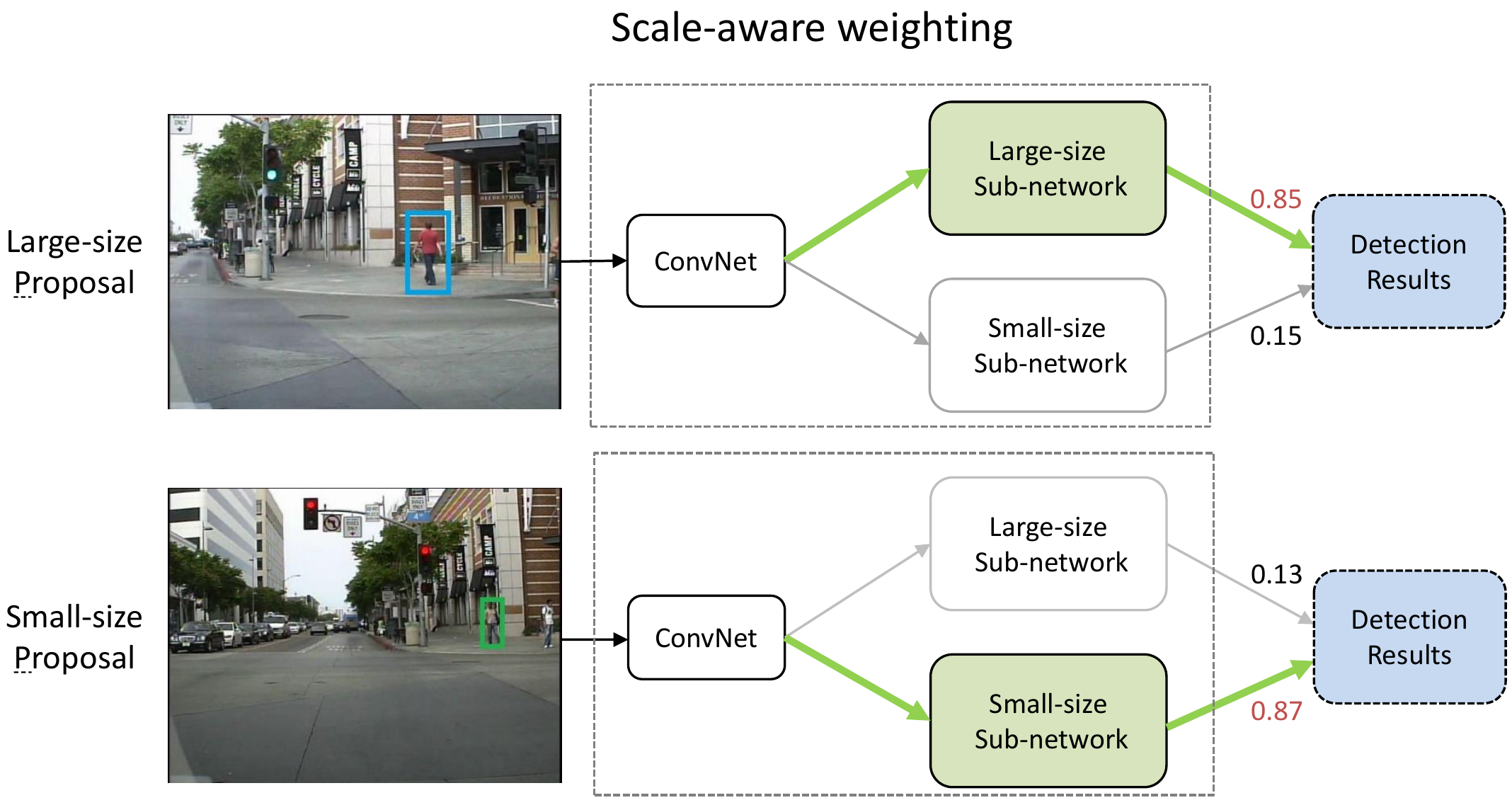}
		\caption{{Illustration of the scale-aware weighting mechanism of our SAF R-CNN. A large-size and a small-size sub-network are learned specifically to detect instances with different sizes. The final result is obtained by fusing the outputs of the two sub-networks according to the object proposal size. Given a large-size object proposal, the weight for the large-size sub-network is high while that for the small-size sub-network is low. In this way, the final result is mainly decided by the large-size network. The situation is the opposite given a small-size object proposal.}}	
		\label{fig:weighting}
	\end{center}
	\vspace{-4mm}
\end{figure*}

Motivated by the above idea, we develop a novel Scale-Aware Fast R-CNN (SAF R-CNN) framework, which is built on the Fast R-CNN pipeline~\cite{girshick2015fast}. The proposed SAF R-CNN integrates a large-size sub-network and a small-size sub-network\footnote{Throughout the paper, we use ``large-size network''/``small-size network'' to refer to a network trained specifically for detecting objects of large/small sizes.} into a unified architecture. As shown in Figure~\ref{fig:weighting}, given an input image with object proposals in it, the SAF R-CNN first passes the raw image through the bottom shared convolutional layers to extract its whole feature maps. Taking these feature maps and the locations of object proposals as inputs, two sub-networks offer different category-level confidence scores and bounding box regressions for each proposal, which are then combined to generate the final detection results using two scale-aware weights predicted by a scale-aware weighting layer that performs a gate function defined over the proposal size. 

SAF R-CNN employs the gate function in a following way to achieve robustness to various scales: it assigns a higher weight for the large-size sub-network when the input has a large size; otherwise, it gives a higher weight  for the small-size sub-network. Such a scale-aware weighting mechanism can be deemed as the soft activations for the two sub-networks, and the final results can always be boosted by the sub-network proper for the current input of certain scales. Therefore, SAF R-CNN can achieve outperforming detection performance in a wide range of input scales. Moreover, since the SAF R-CNN shares convolutional features for the whole image with different object proposals, it is very efficient in terms of both training and testing time.

To sum up, this work makes the following contributions. Firstly, we propose a novel Scale-Aware Fast R-CNN model for pedestrian detection by incorporating a large-size sub-network and a small-size sub-network into a unified architecture following the divide-and-conquer philosophy. Secondly, a scale-aware weighting mechanism is proposed to lift the contribution of the sub-network specialized for the current input scales and boost the final detection performance in a wide input scale range. Thirdly, extensive experiments on several challenging pedestrian datasets demonstrate that SAF R-CNN delivers new state-of-the-art performance on three out of four challenging pedestrian benchmarks.

\section{Related Work}
\textbf{Hand-crafted Model:} The models based on hand-crafted features have been widely used for object detection~\cite{dalal2005histograms,viola2003detecting,wang2009hog,dollar2009integral,compact,dollar2014fast,zhang2014informed,felzenszwalb2010object,zheng2013strip,pang2014distributed,pang2016learning}. Deformable part-based models~\cite{felzenszwalb2010object} consider the appearance of each part and the deformation among parts for detection. The Integral Channel Features (ICF)~\cite{dollar2009integral} and Aggregated Channel Features (ACF)~\cite{dollar2014fast} efficiently extract features such as local sums, histograms, and Haar features using integral images. Wang et al.~\cite{wang2009hog} combined Histograms of Oriented Gradients (HOG) and Local Binary Pattern (LBP) as the feature set to handle partial occlusion. Nam et al.~\cite{nam2014local} introduced an efficient feature transform that removes correlations in local image neighborhoods by extending the features of~\cite{hariharan2012discriminative} to ACF. A multi-order context representation was used in~\cite{chen2013detection} to exploit co-occurrence contexts of different objects. Cai et al.~\cite{compact} combined features of different complexities to seek an optimal trade-off between accuracy and complexity. In addition, some approaches aim to be scale-invariant. Park et al.~\cite{park2010multiresolution} adopted a multi-resolution model that acts as a deformable part-based model when scoring large instances and a rigid template when scoring small instances. Yan et al.~\cite{yan2013robust} proposed to map the pedestrians in different resolutions to a common subspace to reduce the differences of local features. Then a shared detector is learned on the mapped features to distinguish pedestrians from background.

\textbf{Deep Learning Model:} Convolutional neural networks (CNNs) have recently been successfully applied in generic object recognition~\cite{girshick2014rich,he2014spatial,girshick2015fast,ren2015faster,sermanet2013overfeat}. 
Some recent works  focus on improving the performance of pedestrian detection using deep learning methods~\cite{ouyang2013joint,sermanet2013pedestrian,ouyang2012discriminative,ta_cnn}. Sermanet et al.~\cite{sermanet2013pedestrian} used an unsupervised method based on convolutional sparse coding to pre-train CNN for pedestrian detection. Tian et al.~\cite{ta_cnn} jointly optimized pedestrian detection with semantic tasks. In addition, several approaches have also been proposed to improve the scale-invariance of CNN. Gong et al.~\cite{gong2014multi} extracted CNN activations for local patches at three different scales and produced a concatenated feature of the patch by performing orderless VLAD pooling of these activations at each level separately. Xu et al.~\cite{xu2014scale} detected the input pattern at different scales in multiple columns simultaneously and concatenated the top-layer feature maps from all the columns for final classification.  
Previous methods often employ the same filter on the object proposals with various sizes, but the difference of intrinsic characteristics of the large-size and the small-size object proposals have not been fully explored. We explore a simple yet effective framework that consists of a large-size and a small-size sub-network, and fuses their results using the scale-aware weights with respect to the proposal sizes. The two sub-networks are learned specifically to be experts on different input scales, thus achieving high robustness to the scale-variance.

\section{Scale-Aware Fast R-CNN (SAF R-CNN)}

\subsection{Overview of Proposed Model}
The proposed Scale-Aware Fast R-CNN (SAF R-CNN) framework is an ensemble of two scale specific  sub-networks which detect the pedestrians of large  and small sizes, respectively. The detection results of the two sub-networks are then passed  through a  gate function -- defined over the input scale -- for fusion. Such scale-aware collaboration of two sub-networks enables the proposed SAF R-CNN to accurately capture unique characteristics of objects at different scales, and meanwhile the shared convolutional filters in its early layers also incorporate the common characteristics shared by all instances. The architecture of SAF R-CNN is developed based on the popular Fast R-CNN detection framework~\cite{girshick2015fast} due to its superior performance and computation efficiency in detecting general objects. The SAF R-CNN takes the whole image and a number of object proposals as input, and then outputs the detection results. 

\subsection{Pedestrian Proposals Extraction}
In this paper, we utilize the ACF detector~\cite{dollar2014fast} to generate object proposals. The ACF detector is a fast and effective sliding window based detector that performs quite well on rigid object detection. Unlike other proposal methods for detecting generic objects~\cite{uijlings2013selective,arbelaez2014multiscale,zitnick2014edge}, the ACF detector can be trained to detect objects of a specific category, which thus can be used for extracting and mining high-quality object proposals. For fair comparison with the state-of-the-arts~\cite{hosang2015taking,ta_cnn}, we also use the object proposals from ACF detector as inputs. Following the standard setting on the Caltech dataset~\cite{dollar2012pedestrian}, we train an ACF pedestrian detector on the Caltech training set and apply the ACF detector on training and testing images with a low detection threshold of $-70$ to generate object proposals.

\begin{figure*}
	\begin{center}
		\includegraphics[scale=0.50]{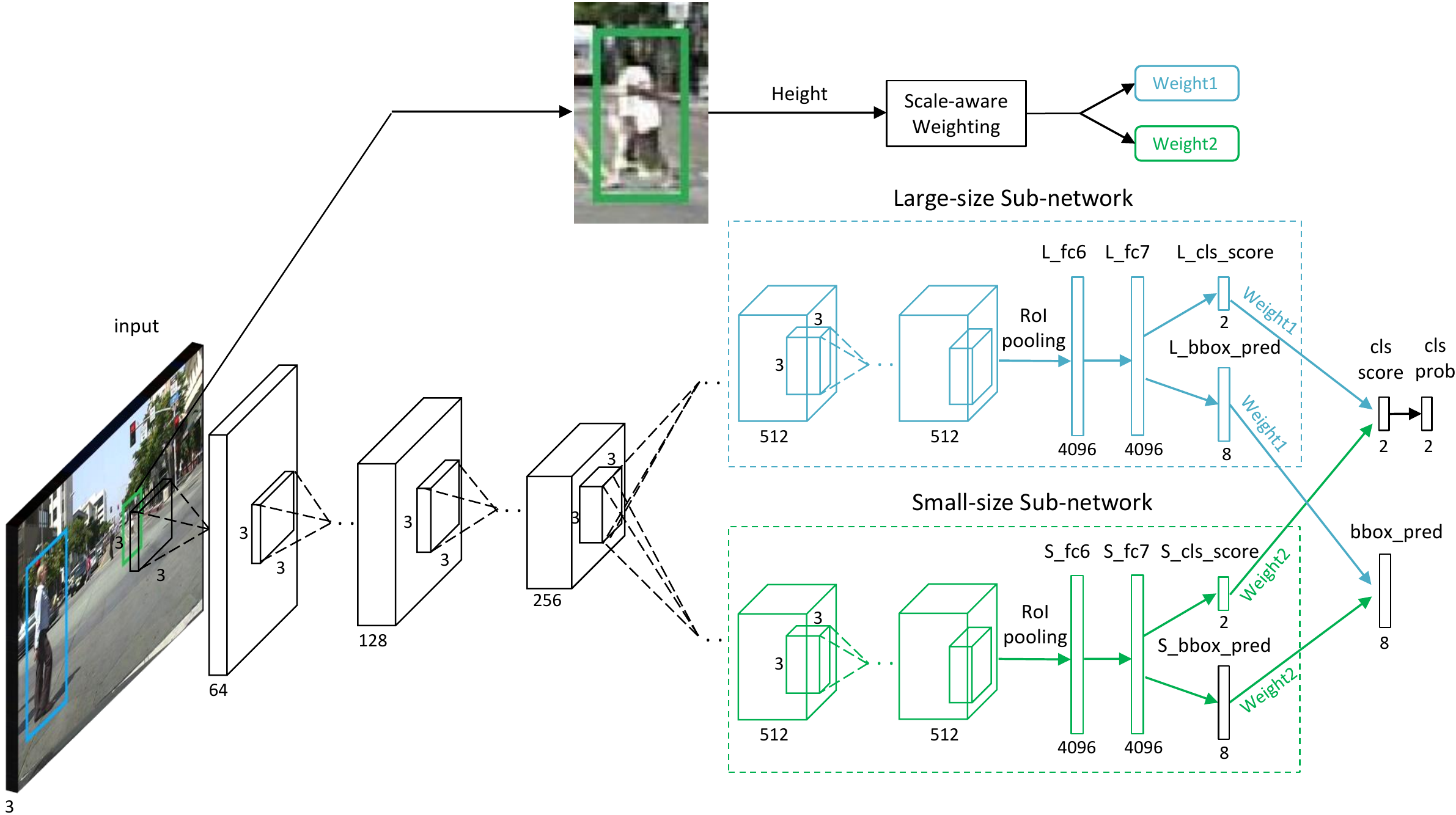}
		\caption{{The architecture of our SAF R-CNN. The features of the whole input image are first extracted by a sequence of convolutional layers and max pooling layers, and then fed into two sub-networks. Each sub-network first utilizes several convolutional layers to further extract scale-specific features. Then, an RoI pooling layer pools the produced feature maps into a fixed-length feature vector and then a sequence of fully connected layers ending up with two output layers are performed to generate scale-specific detection results: one outputs classification scores over $K$ object classes plus a ``background" class and the other outputs refined bounding-box positions for each of the $K$ object classes. Finally, the outputs of the two sub-networks are weighted to obtain the final results with weights from the scale-aware weighting layer which performs a gate function defined over the object proposal sizes.}}	
		\label{fig:architecture}
	\end{center}
	\vspace{-4mm}
\end{figure*}

\subsection{Architecture of SAF R-CNN}
Figure~\ref{fig:architecture} illustrates the architecture of SAF R-CNN in details. The SAF R-CNN passes the input image into several convolutional layers and max pooling layers to extract feature maps. Then the proposed network branches into two sub-networks, which are learned specifically to detect large-size and small-size instances respectively. Each of the two sub-networks takes as input the feature maps produced from the previous convolutional layers, and further extracts features through several convolutional layers to produce feature maps specialized for a specific range of input scales. The region of interest (RoI) pooling layer as proposed in~\cite{girshick2015fast} is utilized to pool the feature maps of each input object proposal into a fixed-length feature vector which is fed into a sequence of fully connected layers.

Each sub-network ends up with two output layers which produce two output vectors per object proposal. Particularly, one layer outputs classification scores over $K$ object classes plus a ``background" class. The other one is the bounding-box regressor which outputs refined bounding-box positions for each of the $K$ object classes. Finally, the outputs from the two sub-networks are weightedly combined to obtain the final result for each input object proposal. Two weights for each object proposal for fusing the outputs from the two sub-networks are given by a scale-aware weighting layer that performs a gate function defined over the object proposal sizes, which will be detailed in Section~\ref{sec:weight}. The classification scores from the two sub-networks are weightedly combined by the computed weights to obtain the final classification score which is fed into a softmax layer to produce softmax probabilities over $K+1$ classes for each input object proposal. Similarly, the bounding box regressions are accordingly combined by the weights to produce the final result for each proposal.

\subsection{Scale-aware Weighting}\label{sec:weight}
As the extracted features from large-size and small-size pedestrians exhibit significant differences, SAF R-CNN incorporates two sub-networks, focusing on the detection of large-size and small-size pedestrians, respectively. A scale-aware weighting layer is designed to perform a gate function which is defined over the sizes of object proposals and used to combine the detection results from two sub-networks. Intuitively, the weights for the two sub-networks should satisfy the following constraints. As illustrated in Figure~\ref{fig:weighting}, given a large-size object proposal, the weight for the large-size network is supposed to be high while that for the small-size sub-network is supposed to be low. The situation is the opposite for a small-size object proposal. Thus by fusing the outputs from the two sub-networks with the weights, SAF R-CNN can be robust to diverse sizes of pedestrian instances. Such a scale-aware weighting mechanism can be deemed as the soft activations for the two sub-networks, and the final results can always be boosted by the proper sub-network for the current input of a certain size.


Note that the size of an object proposal can be measured by either its width or its height. However, for a pedestrian standing with a constant distance to the camera, the height of his bounding box varies little while the width may vary considerably with different poses of the pedestrian. This fact is also described in the Caltech benchmark~\cite{dollar2012pedestrian}. Thus, the height of the bounding box is more stable for measuring the size of a pedestrian. 

In our proposed method, we define a scale-aware weighting layer which performs a gate function over the height of the proposal to adaptively weight the outputs from the two sub-networks. More specifically, let $\boldsymbol{\omega}_l$ and $\boldsymbol{\omega}_s$ denote the output weights computed through the scale-aware weighting layer for the large-size and the small-size sub-network respectively. Given an input object proposal with height $h$, $\boldsymbol{\omega}_l$ is calculated as

\begin{equation}
	\begin{aligned}
		\boldsymbol{\omega}_l = \frac{1}{1 + \alpha \exp^{-\frac{h - \bar{h}}{\beta}}},
		\label{eq:weights_l}
	\end{aligned}
\end{equation}
where $\bar{h}$ denotes the average height of the pedestrians from the training set and $\alpha$ and $\beta$ are two learnable scaling coefficients. We optimize the two parameters via back propagation. The backwards function of the scale-aware weighting layer  computes partial derivative of the loss function $L$ which will be discussed later with respect to $\alpha$ and $\beta$ as

\begin{equation}
	\begin{aligned}
		\frac{\partial L}{\partial \alpha} = - \frac{\exp^{-\frac{h - \bar{h}}{\beta}}}{(1 + \alpha \exp^{-\frac{h - \bar{h}}{\beta}})^{2}} \frac{\partial L}{\partial {\omega}_l} \\
		\frac{\partial L}{\partial \beta} = - \frac{\alpha (h - \bar{h}) \exp^{-\frac{h - \bar{h}}{\beta}}}{\beta^{2} (1 + \alpha \exp^{-\frac{h - \bar{h}}{\beta}})^{2}} \frac{\partial L}{\partial {\omega}_l}.
		\label{eq:derivative_alpha}
	\end{aligned}
\end{equation}


Because the final results are obtained by fusing the outputs from the large-size and the small-size sub-network, we fix the sum of the weights for the two sub-networks as one to avoid improper domination of either model. Thus the weight for the output of the small-size sub-network $\boldsymbol{\omega}_s$ can be simply calculated as $\boldsymbol{\omega}_s = 1 - \boldsymbol{\omega}_l$.

Given a large-size object proposal with a high height, the value of $\boldsymbol{\omega}_l$ goes to 1 while $\boldsymbol{\omega}_s$ is close to 0. Then the final prediction is mainly contributed from the large-size sub-network. On the contrary, given a small-size object proposal with a low height, the final results are mostly determined by the small-size sub-network.

\subsection{Optimization}
Each sub-network in our SAF R-CNN has two sibling output layers. The first sibling layer outputs a discrete confidence score distribution $\boldsymbol{s} = (s_0,...,s_k)$ for each object proposal over $K+1$ categories. The second sibling layer outputs the bounding-box regression offsets for each of the $K$ object classes. The bounding-box regression offsets for the class $k$ can be denoted as $\boldsymbol{t}^k = (t_x^k, t_y^k, t_w^k, t_h^k)$. Following the parameterization scheme in~\cite{girshick2014rich}, $\boldsymbol{t}^k$ specifies the location translation and the bounding box size shift relative to the original location and size of the object proposal. The outputs from the two sub-networks are combined according to the weights computed from the size of the input object proposal as above described. Recall that $\boldsymbol{\omega}_l$ and $\boldsymbol{\omega}_s$ are the weights for the outputs of the large-size and the small-size sub-network respectively. A final predicted discrete confidence score distribution can be computed as

\begin{equation}
	\begin{aligned}
		\boldsymbol{s}_f = \boldsymbol{\omega}_l \times \boldsymbol{s}_l + \boldsymbol{\omega}_s \times \boldsymbol{s}_s,
	\end{aligned}
\end{equation}
where $\boldsymbol{s}_l$ and $\boldsymbol{s}_s$ denote the discrete confidence score distribution output by the first sibling layer of the large-size and the small-size sub-network respectively. Similarly, a final weighted bounding-box regression offset is computed as

\begin{equation}
	\begin{aligned}
		\boldsymbol{t}_f = \boldsymbol{\omega}_l \times \boldsymbol{t}_l + \boldsymbol{\omega}_s \times \boldsymbol{t}_s,
	\end{aligned}
\end{equation}
where $\boldsymbol{t}_l$ and $\boldsymbol{t}_s$ denote the bounding-box regression offsets output by the second sibling layer of the large-size and the small-size sub-network respectively. 

Each training proposal is labeled with a ground-truth class $g$ and a ground-truth bounding-box regression target $t^{*}$. The following multi-task loss $\boldsymbol{L}$ on each object proposal is utilized to jointly train the network parameters of two sub-networks:
\begin{equation}\label{eqn:loss}
\boldsymbol{L} = \boldsymbol{L}_{cls}(s_f,g)+\mathbf{1}[g\geq 1]\boldsymbol{L}_{loc}({t}_f^g, t^{*}),
\end{equation}
where $\boldsymbol{L}_{cls}$ and $\boldsymbol{L}_{loc}$ are the losses for the classification and the bounding-box regression, respectively. In particular, $\boldsymbol{L}_{cls}$ is the log loss and $\boldsymbol{L}_{loc}$ is the smooth $L_1$ loss~\cite{girshick2015fast}. The Iverson bracket indicator function $\mathbf{1}[g\geq 1]$ equals 1 when $g\geq 1$ and 0 otherwise. For background proposals (\emph{i.e. }$g=0$), the $\boldsymbol{L}_{loc}$ is ignored. By jointly training two specialized sub-networks connected by the scale-aware weights with respect to the sizes of object proposals, SAF R-CNN is capable of outputting accurate detection results in a wide range of input scales.



\section{Experiments}
We evaluate the effectiveness of the proposed SAF R-CNN on several popular pedestrian detection datasets including Caltech~\cite{dollar2012pedestrian}, INRIA~\cite{dalal2005histograms}, ETH~\cite{ess2007depth}, and KITTI~\cite{geiger2012we}. More experimental analyses on the effectiveness of each component in our network are further given on the challenging Caltech dataset~\cite{dollar2012pedestrian}.

\subsection{Datasets}
\subsubsection{Caltech}
The Caltech dataset and its associated benchmark~\cite{dollar2012pedestrian} are among the most popular pedestrian detection datasets. It consists of about 10 hours of  videos (30 frames per second) collected from a vehicle driving through urban traffic. Every frame in the raw Caltech dataset has been densely annotated with the bounding boxes of pedestrian instances. There are totally 350,000 bounding boxes of about 2,300 unique pedestrians labeled in 250,000 frames. In the reasonable evaluation setting~\cite{dollar2012pedestrian}, the performance is evaluated on pedestrians over 50 pixels tall with no or partial occlusion. We use dense sampling of the training data (every 4th frame) as adopted in~\cite{compact,nam2014local} for evaluating all variants of SAF R-CNN. During training and testing, the scale of the input image is set as $800$ pixels on the shortest side. 
\subsubsection{INRIA and ETH}
The INRIA pedestrian dataset~\cite{dalal2005histograms} is split into a training and a testing set. The training set consists of 614 positive images and 1,218 negative images. The testing set consists of 288 images. The ETH testing set~\cite{ess2007depth} contains 1,804 images in three video clips. Following the training setting commonly adopted by the best performing approaches~\cite{ouyang2012discriminative}~\cite{felzenszwalb2010object}~\cite{sermanet2013pedestrian}, we train our SAF R-CNN model using the INRIA training set and test it on both the INRIA and the ETH testing sets, in order to evaluate the generalization capacity of our model. We use the 614 positive images in the INRIA training set as training data. Many studies (e.g.,~\cite{krizhevsky2012imagenet}~\cite{ranzato2007unsupervised}) have found that using more training data is beneficial for training deep models. Since the INRIA training set has fewer positive training samples than Caltech training set, we implement Gaussian blurring and motion blurring on the original training images for data augmentation. During training, the scale of the input image is set as $600$ pixels on the shortest side. During testing, we resize the images to make the shortest side as $600$ and $1200$ pixels for the INRIA and the ETH testing set, respectively.
\subsubsection{KITTI}
The challenging KITTI dataset~\cite{geiger2012we} consists of 7,481 training and 7,518 test images, which are captured from an autonomous driving platform. Evaluation is done at three levels of difficulty: easy, moderate and hard, where the difficulty is measured by the minimal scale of the pedestrians to be considered and the occlusion and truncation of the pedestrians. For the moderate setting which is used to rank the competing methods in the benchmark, the pedestrians over 25 pixels tall with no or low partial occlusion and truncation are considered. Since the annotations of the testing set are not available, we split the KITTI training set into train and validation subsets as suggested by~\cite{chen20153d}. The images are resized as $800$ pixels on the shortest side during the training and testing time.  

\subsection{Implementation Details}
We use the pre-trained VGG16 model~\cite{simonyan2014very} to initialize SAF R-CNN, which is used in the most recent state-of-the-art method~\cite{compact}. The first seven convolutional layers and three max pooling layers of the VGG16 network are used as the shared convolutional layers before the two sub-networks to produce feature maps from the entire input image. The rest layers of the VGG16 network are used to initialize both the large-size and the small-size sub-network. The fourth max pooling layer is removed to produce larger feature maps in both sub-networks. We observe that this operation improves the  detection performance. Following Fast R-CNN~\cite{girshick2015fast}, the last max pooling layer of the VGG16 network is replaced by the RoI pooling layer to pool the feature maps of each object proposal into fixed resolution, \emph{i.e.} $7\times7$. The final fully-connected layer and softmax are replaced with two sibling fully-connected layers.

The SAF R-CNN is trained with Stochastic Gradient Descent (SGD) with momentum of 0.9, and weight decay of 0.0005. Each mini-batch consists of 80 randomly sampled object proposals in one randomly selected image, where 20 positive object proposals are with intersection over union (IoU) with the ground truth box larger than 0.5, and the rest 60 object proposals which have IoU with the ground-truth bounding box less than 0.5 act as negative training instances. To compute the scale-aware weights of each proposal for two sub-networks, the initial values of parameter $\alpha$ and $\beta$ in Eqn.~(\ref{eq:weights_l}) are set as $1$ and $10$, respectively.
For data augmentation, images are horizontally flipped with a probability of 0.5.

SAF R-CNN is implemented based on the publicly available Caffe platform~\cite{jia2014caffe}. The whole network is trained on a single NVIDIA GeForce GTX TITAN X GPU with 12GB memory. For training, the first four convolutional layers in the network keep constant parameters initialized from the pre-trained VGG16 model. The other layers update parameters with an initial learning rate of 0.001 which is lowered to 1/10 of the current rate after every 4 epochs. We fine-tune the networks for about 7 epochs on the training set. 

\subsection{Comparison with State-of-the-arts}
\subsubsection{Caltech}
\begin{figure}
	\begin{center}
		\includegraphics[scale=0.50]{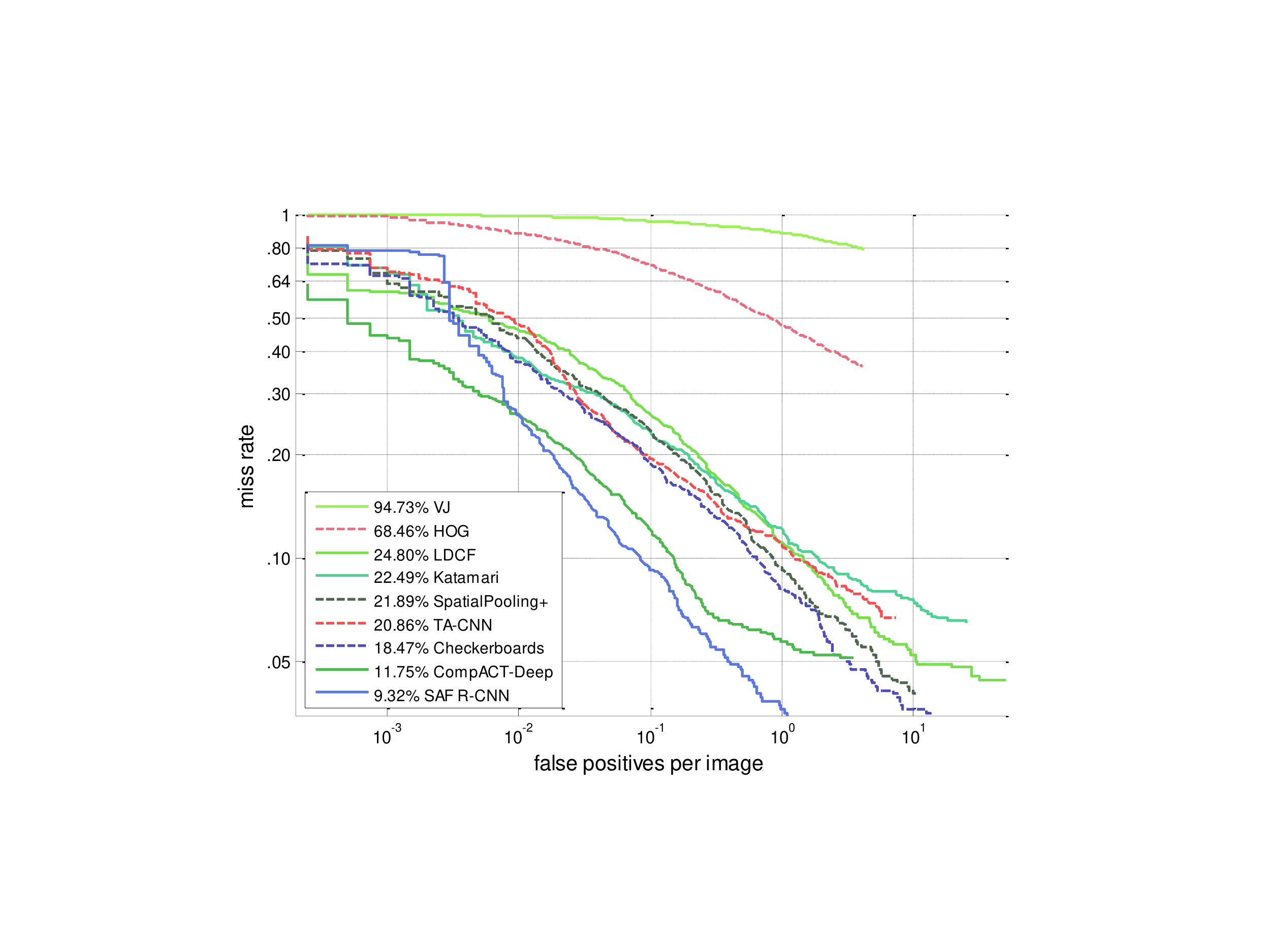}
		\caption{{The comparison of SAF R-CNN on pedestrian detection with all recent state-of-the-art methods on the Caltech dataset. The SAF R-CNN outperforms other methods with the lowest log-average miss rate of 9.32$\%$.}}
		\label{fig:Caltech_Overall_Results}
	\end{center}
	\vspace{-4mm}
\end{figure}

We use the Caltech training set to train our model and evaluate it on the Caltech testing set. The overall experimental results are reported in Figure~\ref{fig:Caltech_Overall_Results}. We compare the result of SAF R-CNN
with all the existing methods that achieved best performance on the Caltech testing set, including VJ~\cite{viola2004robust}, HOG~\cite{dalal2005histograms}, LDCF~\cite{nam2014local}, Katamari~\cite{benenson2014ten}, SpatialPooling+~\cite{paisitkriangkrai2014strengthening}, TA-CNN~\cite{ta_cnn}, Checkerboards~\cite{zhang2015filtered}, and CompACT-Deep~\cite{compact}. It can be observed that SAF R-CNN outperforms other methods by a large margin and achieves the lowest log-average miss rate of 9.32$\%$, which is significantly lower than the current state-of-the-art approach CompACT-Deep~\cite{compact}, by 2.43$\%$.

\subsubsection{INRIA and ETH}
\begin{figure}
	\begin{center}
		\includegraphics[scale=0.45]{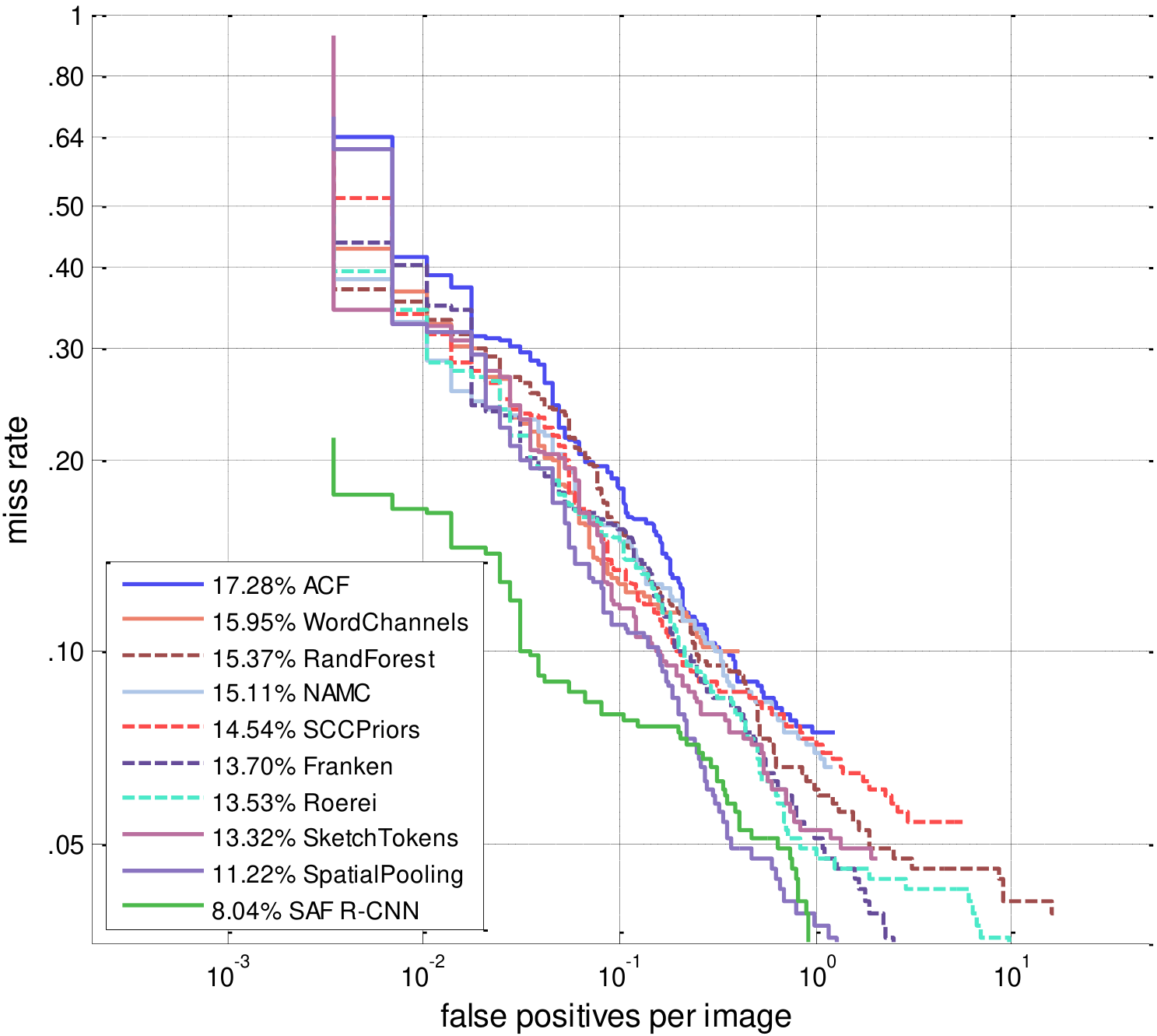}
		\caption{{The comparison of SAF R-CNN on pedestrian detection with all recent state-of-the-art methods on the INRIA dataset. The SAF R-CNN outperforms other methods with the lowest log-average miss rate of $8.04\%$.}}
		\label{fig:INRIA_Overall_Results}
	\end{center}
	\vspace{-4mm}
\end{figure}

\begin{figure}
	\begin{center}
		\includegraphics[scale=0.55]{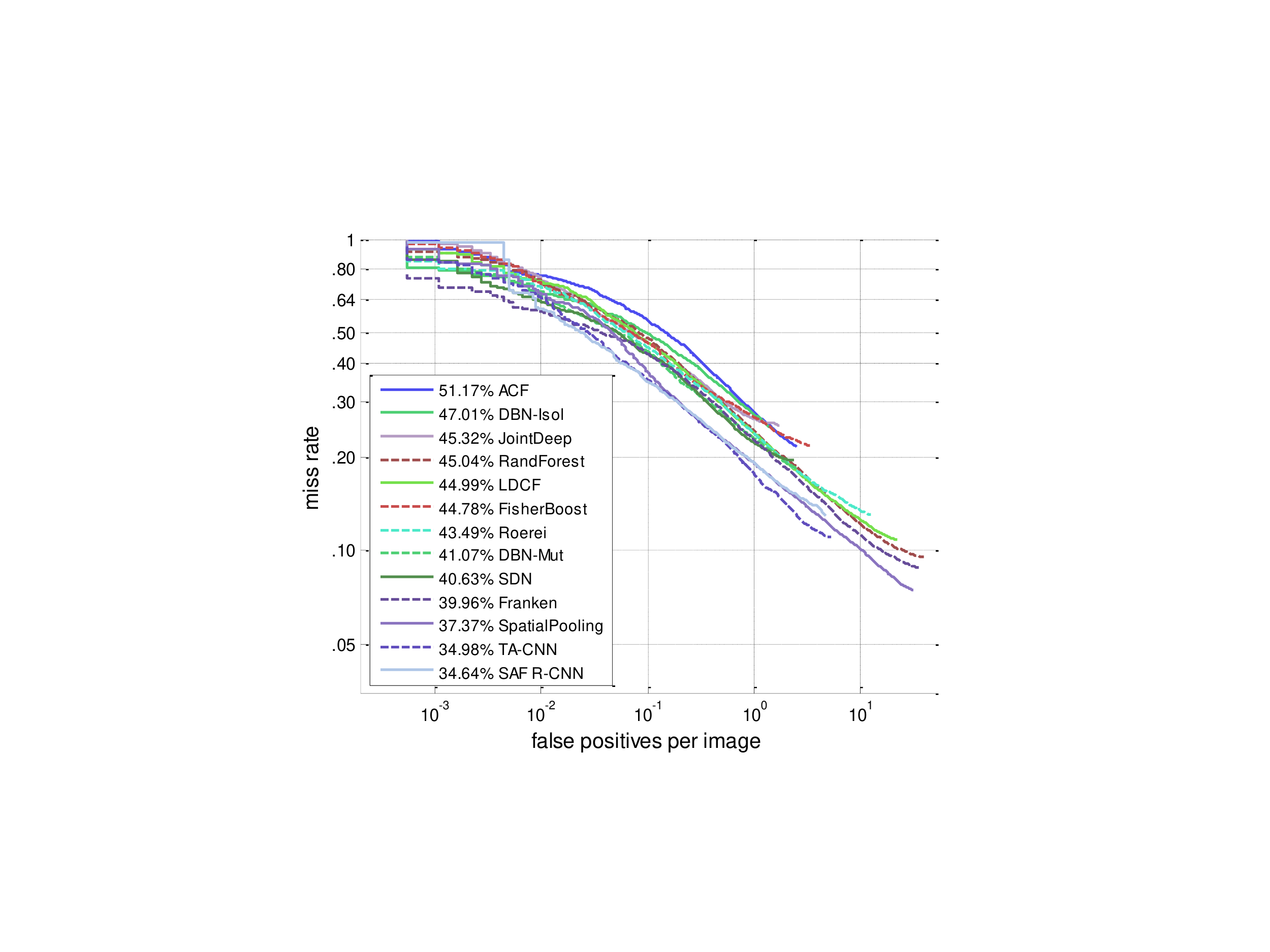}
		\caption{{The comparison of SAF R-CNN on pedestrian detection with all recent state-of-the-art methods on the ETH dataset. The SAF R-CNN outperforms other methods with the lowest log-average miss rate of $34.64\%$.}}
		\label{fig:ETH_Overall_Results}
	\end{center}
	\vspace{-4mm}
\end{figure}

We train our model using the INRIA training set and evaluate the model on both INRIA and ETH testing sets. Figure~\ref{fig:INRIA_Overall_Results} and Figure~\ref{fig:ETH_Overall_Results} provide the comparisons of the proposed method with several best-performing methods~\cite{paisitkriangkrai2014strengthening}~\cite{ta_cnn}~\cite{dollar2014fast}. It can be observed that our model achieves the lowest miss rate on both datasets. For the INRIA dataset, our method obtains the miss rate of $8.04\%$, which outperforms the second best method~\cite{paisitkriangkrai2014strengthening} by $3.18\%$. For the ETH dataset, the miss rate of our model is $34.64\%$ compared with $34.98\%$ of Tian et al.~\cite{ta_cnn} and $37.37\%$ of Paisitkriangkrai et al.~\cite{paisitkriangkrai2014strengthening}. In general, the proposed method outperforms other best-performing methods and achieves state-of-the-art performance on both datasets, which not only validates its superiority in accurate pedestrian detection after tuning in scene (INRIA), but also verifies its generalization capacity to other scenarios (ETH).

\subsubsection{KITTI}
\begin{figure}
	\begin{center}
		\includegraphics[scale=0.50]{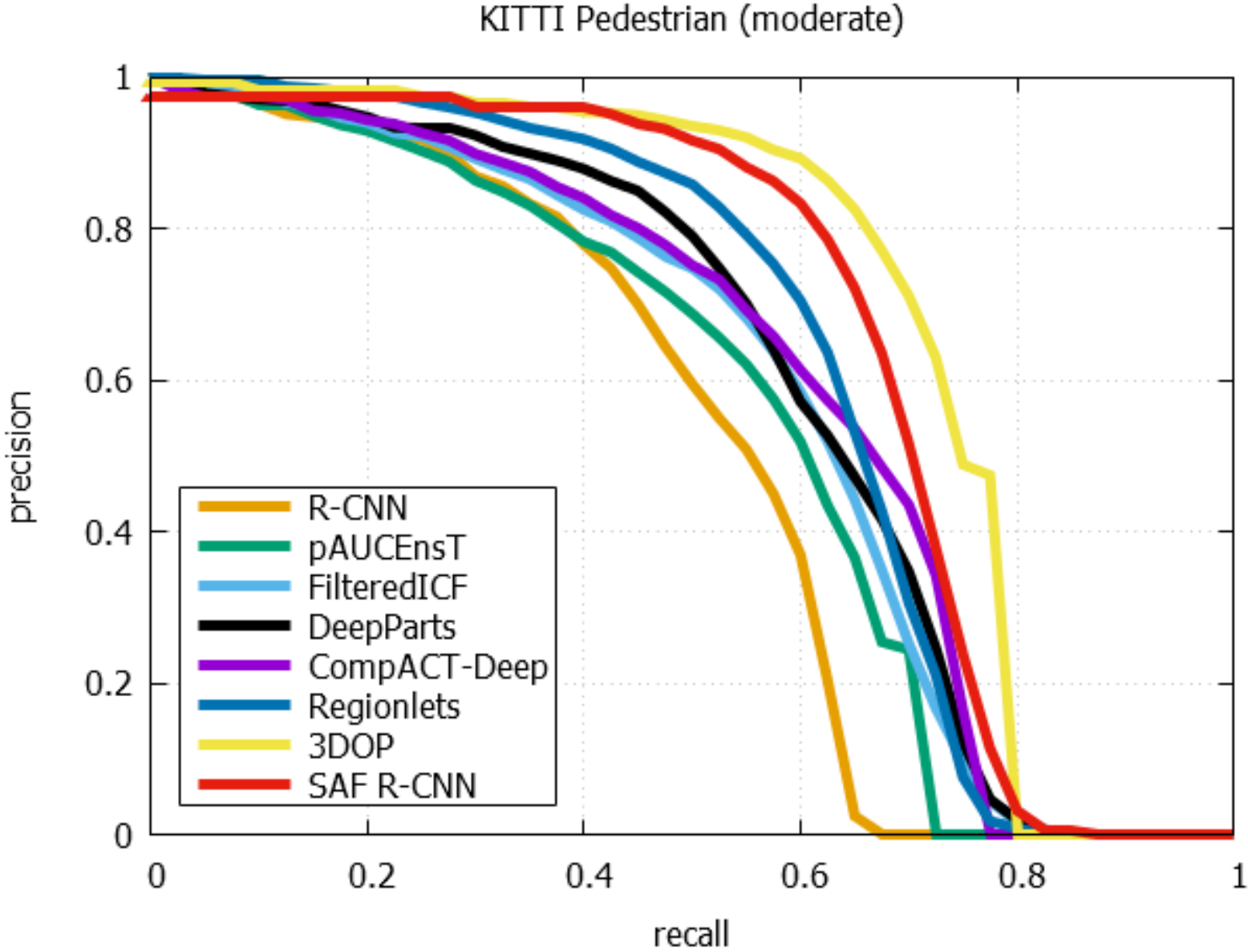}
		\caption{The comparison of SAF R-CNN on pedestrian detection with recent state-of-the-art methods on the KITTI dataset (moderate).}	
		\label{fig:Kitti_Results}
	\end{center}
	\vspace{-4mm}
\end{figure}

We train our model using the KITTI training set and evaluate the model on the testing set of the KITTI benchmark. The detection results and performance comparisons of the proposed method with several best-performing methods ~\cite{hosang2015taking}~\cite{paisitkriangkrai2014strengthening}~\cite{zhang2015filtered}~\cite{tian2015deep}~\cite{compact}~\cite{wang2013regionlets}~\cite{chen20153d} are presented in Table~\ref{tab:Kitti_Results} and Figure~\ref{fig:Kitti_Results}. It can be observed that our model achieves promising results, i.e., $77.93\%$, $65.01\%$, and $60.42\%$ in terms of AP on easy, moderate, and hard subsets respectively, which outperforms most of the previous methods tested on this benchmark by a large margin. Overall, the approach competitive with our model is the 3DOP method of~\cite{chen20153d}. However, this work adopts stereo information by using the left and right images of the KITTI training set while our model is trained with only the left images.


\begin{table}\setlength{\tabcolsep}{3pt}
	\centering\scriptsize
	\caption{Average Precision (AP) (in \%) on the testing set of the KITTI Dataset.}\label{tab:Kitti_Results}
	\renewcommand{\arraystretch}{1.3}
	\begin{tabular}{c | c | c | c}
		\hline
		{\bf Methods} & {\bf Easy} & {\bf Moderate} & {\bf Hard}\\ \hline
		R-CNN & 61.61 & 50.13 & 44.79 \\
		pAUCEnsT & 65.26 & 54.49 & 48.60 \\
		FilteredICF & 67.65 & 56.75 & 51.12 \\
		DeepParts & 70.49 & 58.67 & 52.78 \\
		CompACT-Deep & 70.69 & 58.74 & 52.71 \\
		Regionlets & 73.14 & 61.15 & 55.21 \\		
		3DOP & 81.78 & 67.47 & 64.70 \\	\hline			
		SAF R-CNN & 77.93 & 65.01 & 60.42 \\				
		\hline
	\end{tabular}%
\end{table}%

\subsection{Ablations Studies}
This subsection is devoted to investigating the effectiveness of different components of SAF R-CNN. The performance achieved by different variants of the SAF R-CNN and parameter settings are reported in the following. All experiments are performed on the challenging Caltech dataset.

\subsubsection{Feature Map Size}
\begin{figure}
	\begin{center}
		\includegraphics[scale=0.58]{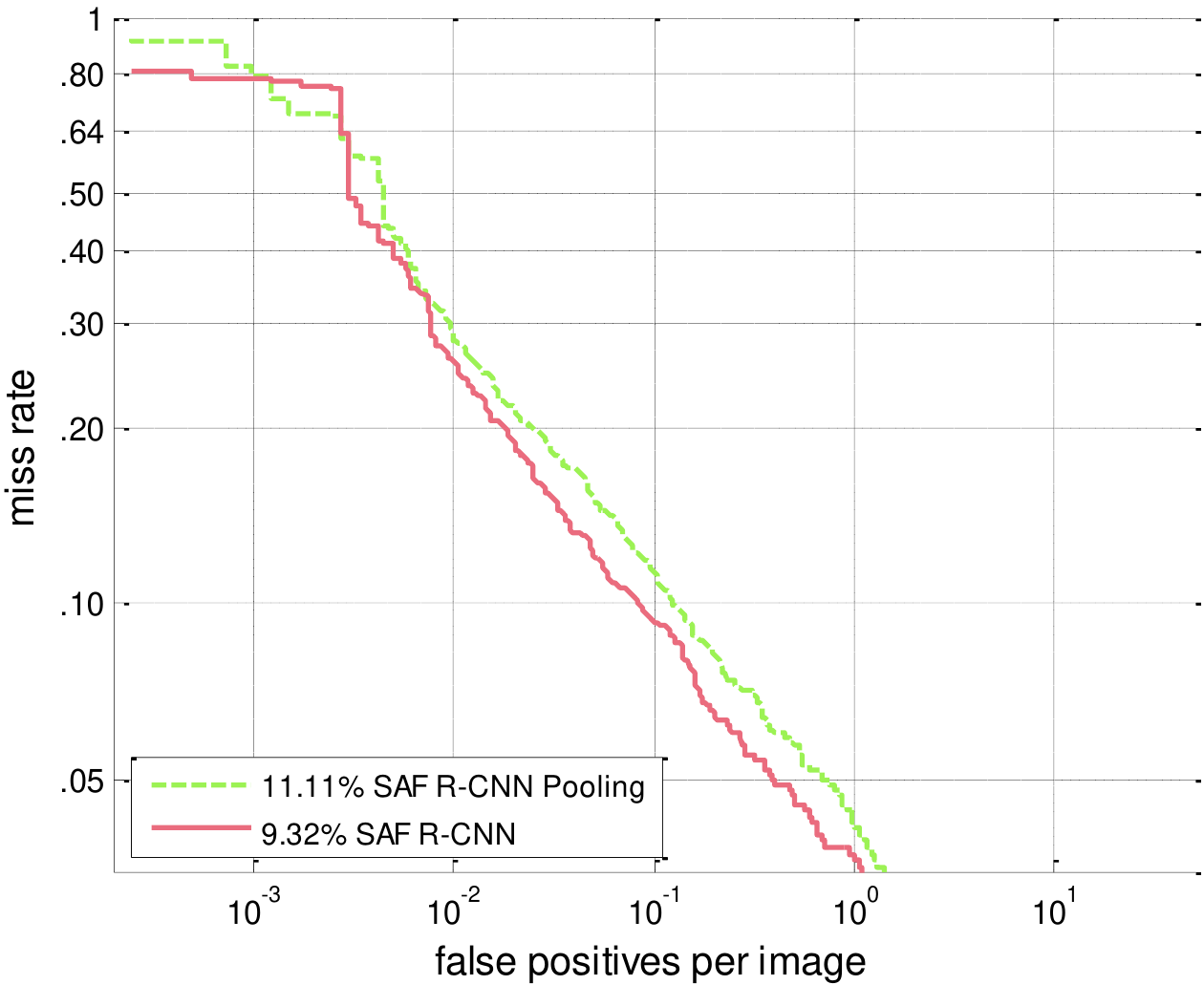}
		\caption{{The comparison of using different sizes of feature maps in our SAF R-CNN. The result of SAF R-CNN is compared with its variant in which the fourth max pooling layer is preserved in the sub-networks, in order to demonstrate the effectiveness of using larger feature maps to detect small pedestrian instances.}}	
		\label{fig:Caltech_Pooling}
	\end{center}
	\vspace{-4mm}
\end{figure}

To generate larger feature maps for small-size object proposals, we remove the fourth max pooling layer in the VGG16 model for training the large-size and the small-size sub-network. To analyze the effectiveness of this strategy, the results of one variant of SAF R-CNN that preserves the fourth max pooling layer in both sub-networks are reported, \emph{i.e.} ``SAF R-CNN Pooling" in Figure~\ref{fig:Caltech_Pooling}. Compared to ``SAF R-CNN", the smaller feature maps for the input object proposals are generated by ``SAF R-CNN Pooling". It can be observed that the SAF R-CNN decreases the miss rate by 1.79$\%$ compared to ``SAF R-CNN Pooling", verifying that larger feature maps are beneficial for the performance improvement in detecting small pedestrian instances. 

\subsubsection{Shared Convolutional Layers}
\begin{figure}
	\begin{center}
		\includegraphics[scale=0.58]{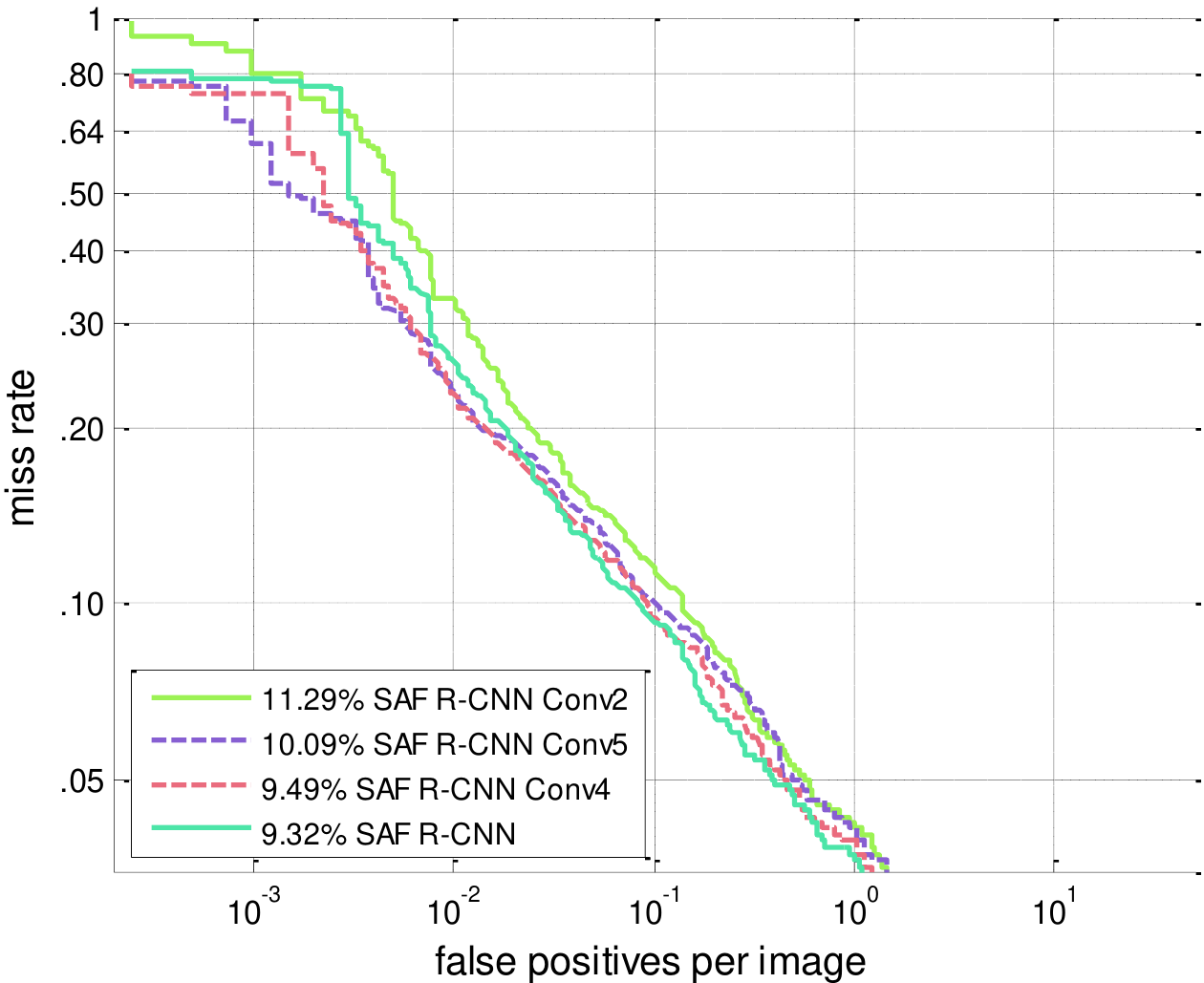}
		\caption{{The comparison of using different convolutional layers as the shared convolutional layers before the two sub-networks. The result of SAF R-CNN is compared with the variant in which the first four convolutional layers and two max pooling layers act as the shared convolutional layers, denoted as ``SAF R-CNN Conv2", the variant where the first ten convolutional layers and three max pooling layers are used as the shared convolutional layers, denoted as ``SAF R-CNN Conv4", and the variant where shared features are extracted from the first thirteen convolutional layers and three max pooling layers, denoted as ``SAF R-CNN Conv5".}}	
		\label{fig:Caltech_Shared_Features}
	\end{center}
	\vspace{-4mm}
\end{figure}

In the SAF R-CNN, we use the first seven convolutional layers and three max pooling layers of the VGG16 network as the shared convolutional layers before the two sub-networks to extract feature maps from the entire input image. To verify the advantage of using these convolutional layers to generate shared features, we evaluate the performance of the variants where shared features are produced by different convolutional layers of the VGG16 network. In Figure~\ref{fig:Caltech_Shared_Features}, ``SAF R-CNN Conv2" denotes the variant in which the first four convolutional layers and two max pooling layers are used as the shared convolutional layers before the two sub-networks. ``SAF R-CNN Conv4" denotes the variant in which the first ten convolutional layers and three max pooling layers act as the shared convolutional layers. ``SAF R-CNN Conv5" represents the variant where shared features are extracted from the first thirteen convolutional layers and three max pooling layers. Compared with ``SAF R-CNN Conv2", ``SAF R-CNN Conv4", and ``SAF R-CNN Conv5", SAF R-CNN improves the performance by $1.97\%$, $0.17\%$, and $0.77\%$, respectively, which verifies that better shared features can be provided for the two sub-networks using the first seven convolutional layers and three max pooling layers of the VGG16 network.

\subsubsection{Scale-aware Weighting}
\begin{figure}
	\begin{center}
		\includegraphics[scale=0.55]{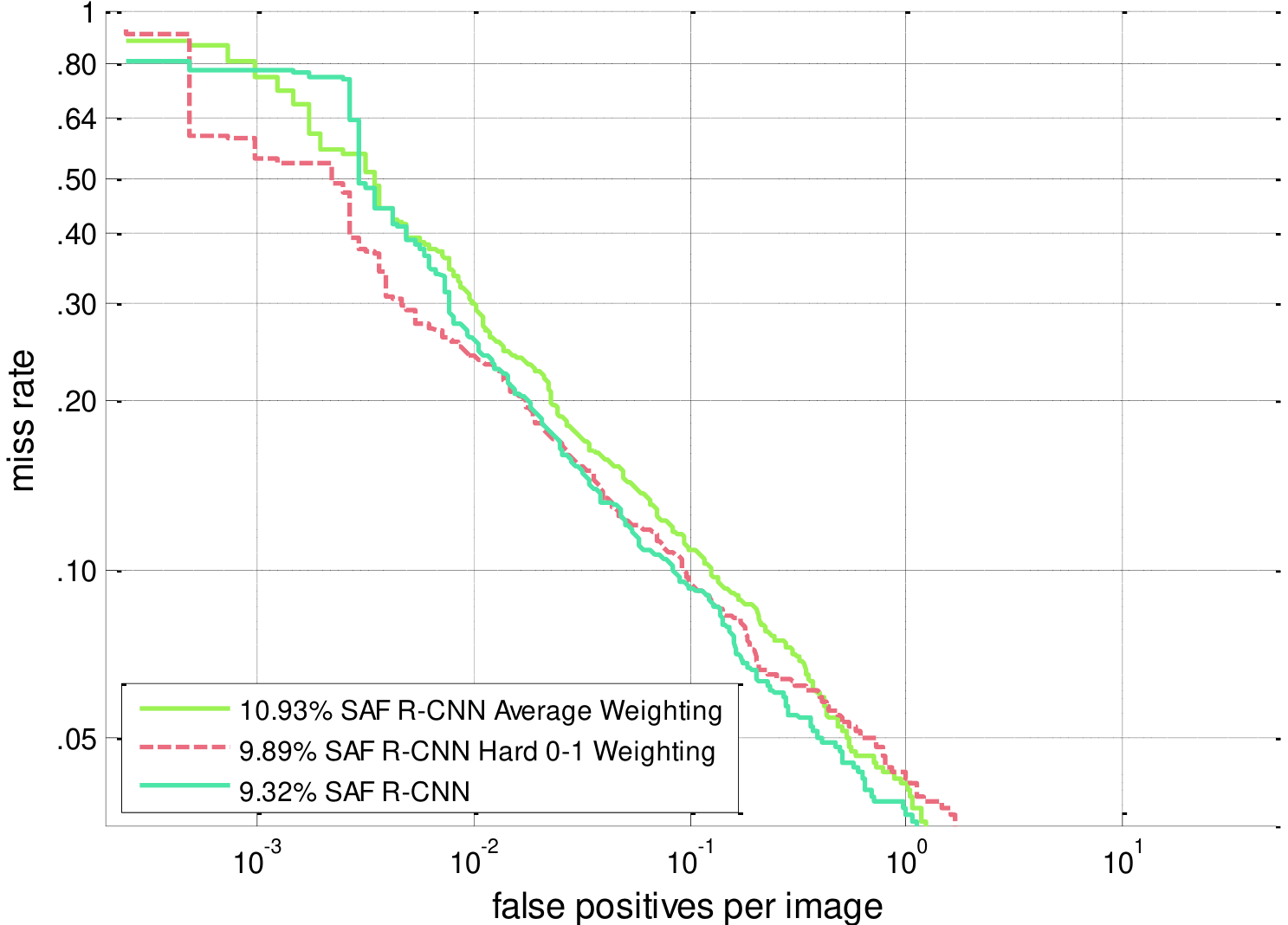}
		\caption{{The comparison of scale-aware weighting with averaging and hard 0-1 weighting strategies. The SAF R-CNN adaptively combines the outputs of the two sub-networks using the scale-aware weighing strategy. To verify its effectiveness, the result of SAF R-CNN is compared with the network which obtains the final results by averaging the outputs of the two sub-networks, denoted as ``SAF R-CNN Average Weighting", and the network which assigns hard 0-1 weights to the two sub-networks, denoted as ``SAF R-CNN Hard 0-1 Weighting".}}	
		\label{fig:Caltech_Weighting}
	\end{center}
	\vspace{-4mm}
\end{figure}

In the SAF R-CNN, the weights for combining the outputs of two sub-networks are computed by the scale-aware weighting layer which performs the gate function defined over the input height of the object proposal. To analyze the effectiveness of our scale-aware weighting strategy, the variant ``SAF R-CNN Average Weighting" is compared, which combines the outputs from the two sub-networks using the equal weights (\emph{i.e.} $0.5$). We also compare SAF R-CNN with the variant ``SAF R-CNN Hard 0-1 Weighting", which assigns hard 0-1 weights to the two sub-networks according to the height of the object proposal. This is equivalent to the variant using the gate function with very small $\beta$. Both ``SAF R-CNN Average Weighting" and ``SAF R-CNN Hard 0-1 Weighting" follow the same fine-tuning step as SAF R-CNN. Figure~\ref{fig:Caltech_Weighting} shows that although these networks are with the same network architecture, the SAF R-CNN decreases the miss rate by $1.61\%$ compared to the ``SAF R-CNN Average Weighting" and $0.57\%$ compared to the ``SAF R-CNN Hard 0-1 Weighting". It demonstrates that adaptively weighting the outputs of the two sub-networks according to the object proposal size is beneficial for the final performance improvement, which makes SAF R-CNN robust to various sizes of the pedestrian instances.


\subsubsection{Input Image Scale} 
\begin{figure}
	\begin{center}
		\includegraphics[scale=0.58]{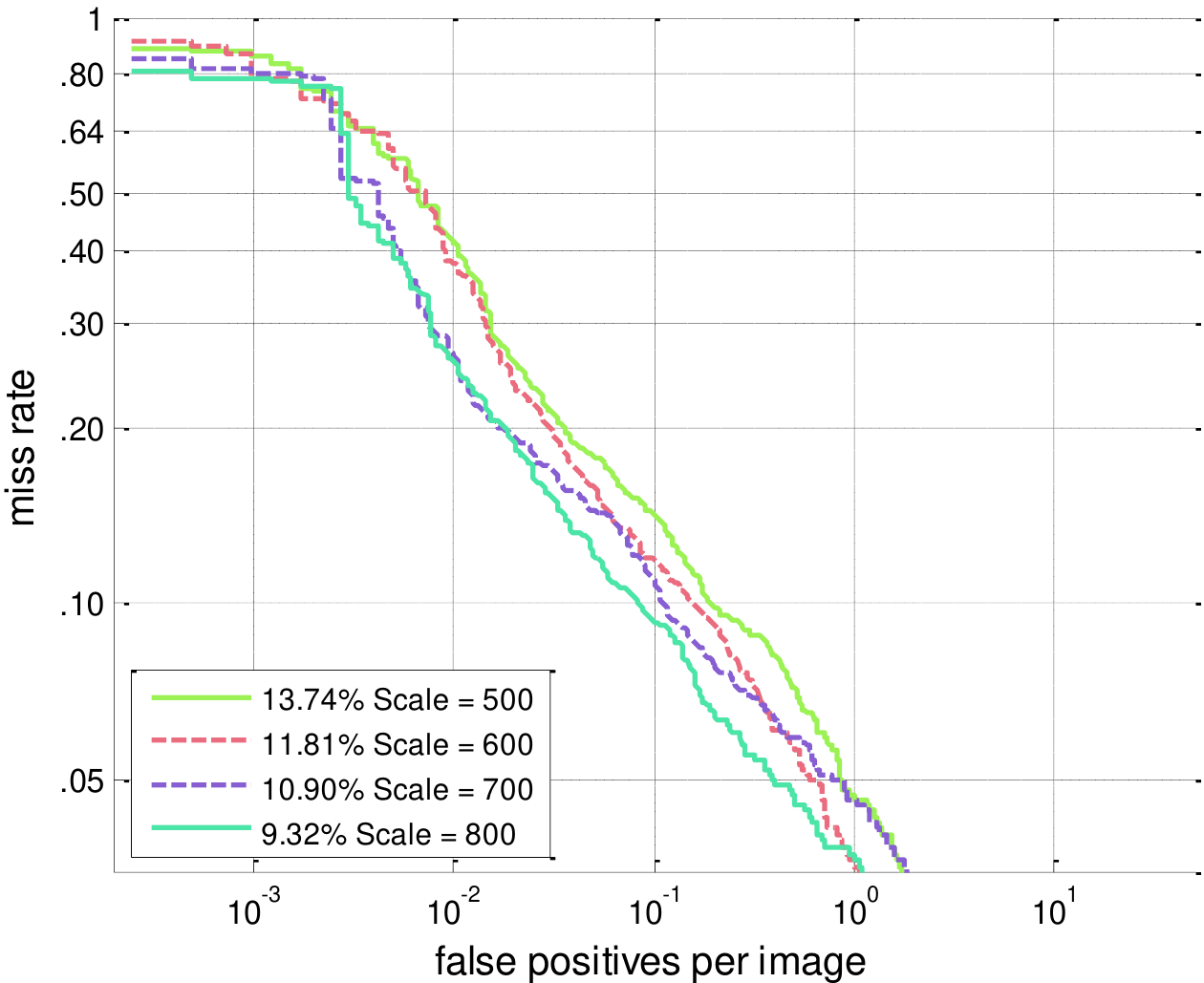}
		\caption{{Performance comparison of using different scales of the input image as the input of SAF R-CNN.}}	
		\label{fig:Caltech_Scales}
	\end{center}
	\vspace{-4mm}
\end{figure}

Several experiments have been conducted to investigate the effect of the input image scale on the detection performance. The results for SAF R-CNN with input image scales of 500, 600, 700 and 800 are shown in  Figure~\ref{fig:Caltech_Scales}. It can be observed that the miss rate decreases along with the increase of the input image scale. This verifies that a larger scale of the whole image can help produce more informative feature maps for each proposal to assist in detecting small-size pedestrian instances. From our experiments, only minor improvement is observed when using larger image scales (such as 1000) but higher computation complexity is required. To balance the computation cost and detection accuracy, we select $800$ as the scale of the input image in all other experiments. 

\begin{figure*}
	\begin{center}
		\includegraphics[scale=0.75]{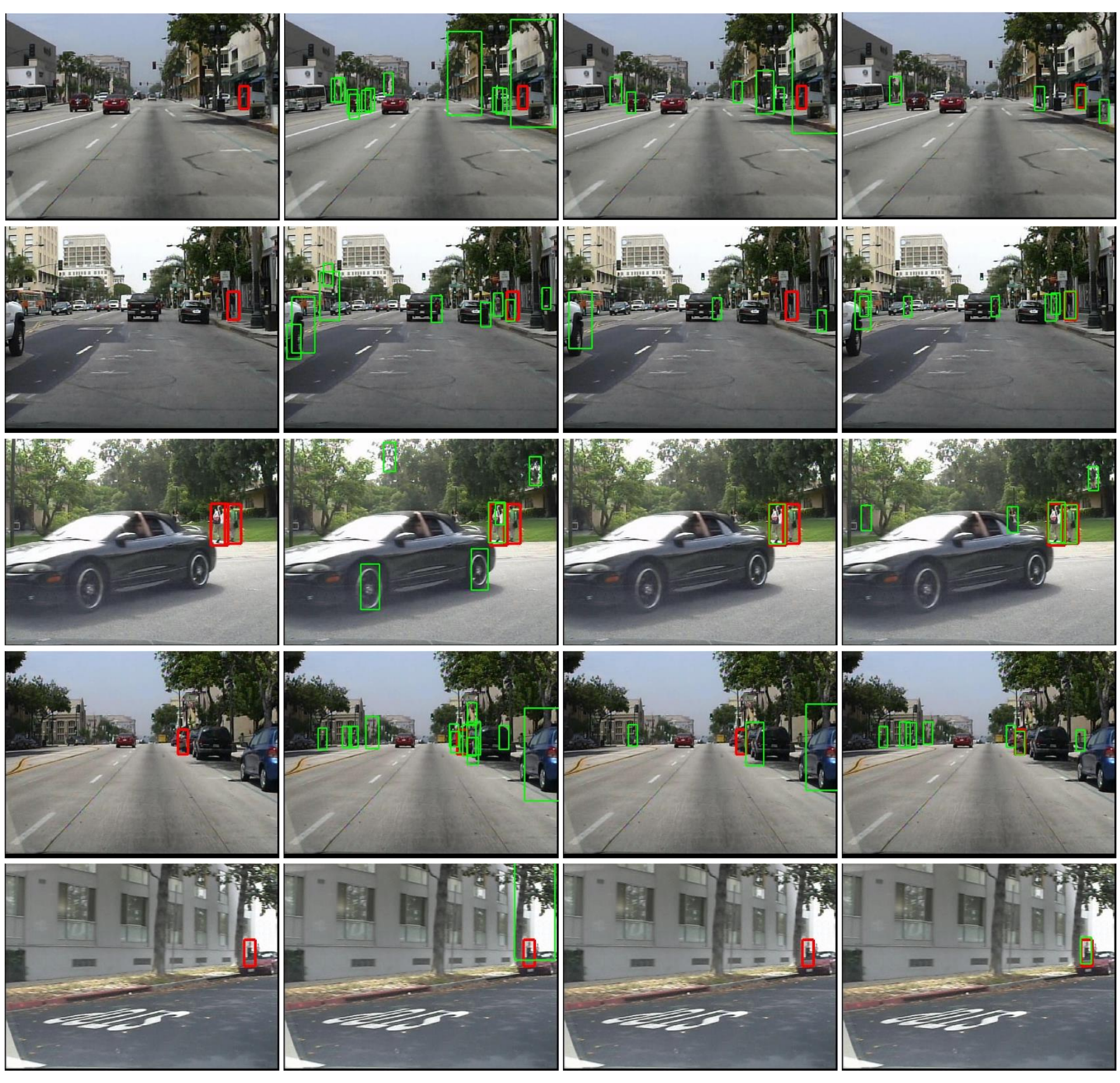}
		\caption{{Comparison of pedestrian detection results with other state-of-the-art methods. The first column shows the input images with ground-truths annotated with red rectangles. The rest columns show the detection results (green rectangles) of TA-CNN~\cite{ta_cnn}, CompACT-Deep~\cite{compact} and SAF R-CNN respectively. Our SAF R-CNN can successfully detect most small-size instances which the other two state-of-the-art methods have missed. For better viewing, please see original PDF file.}}	
		\label{fig:visualization}
	\end{center}
	\vspace{-4mm}
\end{figure*}

\subsubsection{Comparisons with R-CNN, Fast R-CNN, and Faster R-CNN}
We also compare SAF R-CNN with R-CNN~\cite{girshick2014rich}, Fast R-CNN~\cite{girshick2015fast}, and Faster R-CNN~\cite{ren2015faster} for pedestrian detection, shown in Table~\ref{tab:fast_rcnn}. R-CNN~\cite{girshick2014rich} addresses the scale-variance problem by resizing the proposals into a fixed image scale while Fast R-CNN~\cite{girshick2015fast} uses two ways to deal with the scale problem. One is the brute-force approach in which the input image is resized into a pre-defined size on the shortest side, denoted as ``Fast R-CNN single-scale". For fair comparison with our SAF R-CNN, the input image of ``Fast R-CNN single-scale" is resized to 800 pixels on the shortest side. The other one is the multi-scale approach which utilizes multi-scale image pyramids for each image, denoted as ``Fast R-CNN multi-scale". The same five scales of 480, 576, 688, 864 and 1200 are adopted to construct the input image pyramid as specified in ~\cite{he2014spatial}. For Faster R-CNN, we resize the input image to 800 pixels on the shortest side to make fair comparison with SAF R-CNN. It can be seen that SAF R-CNN significantly outperforms all four baselines. It verifies the superiority of using our scale-aware weighting technique in SAF R-CNN to detect the pedestrian instances with various sizes. We also compare the testing time of SAF R-CNN with other baselines. SAF R-CNN is only a little slower than ``Fast R-CNN single-scale" and Faster R-CNN, and 9.0$\times$ faster than R-CNN and 5.2$\times$ faster than ``Fast R-CNN multi-scale". This observation further demonstrates the advantage of SAF R-CNN which yields a large improvement in miss rate with low computation cost.

\begin{table}\setlength{\tabcolsep}{3pt}
	\centering\scriptsize
	\caption{Comparison of the miss rates and testing time with other four methods for scale-variance problem, including ``R-CNN"~\cite{girshick2014rich}, ``Fast R-CNN single-scale"~\cite{girshick2015fast} shorten as ``Fast R-CNN-S", ``Fast R-CNN multi-scale"~\cite{girshick2015fast} shorten as ``Fast R-CNN-M", and ``Faster R-CNN"~\cite{ren2015faster}.}\label{tab:fast_rcnn}
	\renewcommand{\arraystretch}{1.3}
	\begin{tabular}{c|c|c|c|c|cccccccccccccccccc}
		\hline
		\textbf{Method} &\textbf{R-CNN} &\textbf{Fast R-CNN-S} &\textbf{Fast R-CNN-M} &\textbf{Faster R-CNN} &\textbf{Ours} \\
		\hline
		\textbf{miss rate (\%)} & 12.77 & 13.70 & 11.67 & 17.60 & 9.32 \\
		\textbf{test rate (s/im)}  & 5.31 & 0.34 & 3.04 & 0.22 & 0.59 \\
		\hline
	\end{tabular}%
\end{table}%

\subsubsection{Visualization of Detection Results}
Several detection results of our SAF R-CNN, TA-CNN~\cite{ta_cnn} and CompACT-Deep~\cite{compact} are visualized in Figure~\ref{fig:visualization} to further demonstrate the superiority of SAF R-CNN in detecting small-size instances. The first column shows the input images and the rest three columns sequentially show the detection results by TA-CNN~\cite{ta_cnn}, CompACT-Deep~\cite{compact} and our SAF R-CNN. The ground-truth bounding boxes of pedestrians are annotated with red rectangles, and the green rectangles represent the detected instances by our SAF R-CNN and the two baselines. One can observe that SAF R-CNN can successfully detect most of the small-size pedestrian instances that TA-CNN and CompACT-Deep have missed, especially for those with obscured boundaries. The last row in Figure~\ref{fig:visualization} shows that our SAF R-CNN is robust to heavy occlusion of pedestrians and large background clutters.

\section{Conclusion and Future work} 
In this paper, we proposed a novel Scale-Aware Fast R-CNN (SAF R-CNN) model which incorporates a large-size sub-network and a small-size sub-network into a unified architecture to deal with various sizes of pedestrian instances in the image.  By sharing convolutional filters in early layers for extracting common features and combining the outputs of the two sub-networks using the designed scale-aware weighing mechanism, SAF R-CNN is capable of training the specialized sub-networks for large-size and small-size pedestrian instances in order to capture their unique characteristics. Extensive experiments have demonstrated that the proposed SAF R-CNN is superior in detecting small-size pedestrian instances and achieves state-of-the-art performance on several challenging benchmarks. In future, we will extend the proposed SAF R-CNN to general object detection.

\vspace{-0.05in}
\bibliographystyle{plain}
\scriptsize
\bibliography{tip}

\end{document}